\documentclass{article}

\usepackage{PRIMEarxiv}
\usepackage{bm}
\usepackage{multirow}
\usepackage[utf8]{inputenc} 
\usepackage[T1]{fontenc}    
\usepackage{hyperref}       
\usepackage{url}            
\usepackage{booktabs}       
\usepackage{amsfonts}       
\usepackage{nicefrac}       
\usepackage{microtype}      
\usepackage{lipsum}
\usepackage{fancyhdr}       
\usepackage{graphicx}       
\graphicspath{{media/}} 
\usepackage{amsmath}
\usepackage{ulem}
\usepackage{xcolor} 
\usepackage{bm}
\usepackage{wrapfig}
\usepackage{multirow}
\definecolor{darkred}{rgb}{0.6, 0.15, 0.15}

\hypersetup{
    colorlinks=True, 
    pdfborder={0 0 0} 
}
\title{WormKAN: Are KAN Effective for Identifying and Tracking Concept Drift in Time Series?

}

\author{
  Kunpeng Xu, Lifei Chen, Shengrui Wang \\
  Université de Sherbrooke \\
  Québec, Canada
}

\begin{document}
\maketitle

\begin{abstract}
Dynamic concepts in time series are crucial for understanding complex systems such as financial markets, healthcare, and online activity logs. These concepts help reveal structures and behaviors in sequential data for better decision-making and forecasting. However, existing models often struggle to detect and track concept drift due to limitations in interpretability and adaptability. To address this challenge, inspired by the flexibility of the recent Kolmogorov-Arnold Network (KAN), we propose WormKAN, a concept-aware KAN-based model to address concept drift in co-evolving time series. WormKAN consists of three key components: Patch Normalization, Temporal Representation Module, and Concept Dynamics. Patch normalization processes co-evolving time series into patches, treating them as fundamental modeling units to capture local dependencies while ensuring consistent scaling. The temporal representation module learns robust latent representations by leveraging a KAN-based autoencoder, complemented by a smoothness constraint, to uncover inter-patch correlations. Concept dynamics identifies and tracks dynamic transitions, revealing structural shifts in the time series through concept identification and drift detection. These transitions, akin to passing through a \textit{wormhole}, are identified by abrupt changes in the latent space. Experiments show that KAN and KAN-based models (WormKAN) effectively segment time series into meaningful concepts, enhancing the identification and tracking of concept drift.
\end{abstract}


\section{Introduction}

Time series analysis plays a crucial role in various fields such as finance, healthcare, and meteorology. Recently, deep learning models have made significant strides in forecasting tasks. Notable advancements include Informer \cite{zhou2021informer}, which introduces an efficient transformer-based approach, and TimeMixer \cite{wang2024timemixer}, which employs a multiscale mixing approach. N-BEATS \cite{oreshkin2019n} and OneNet \cite{wen2024onenet}, the ensemble deep learning models, demonstrate strong forecasting performance. Additionally, large language models (LLMs) like UniTime \cite{liu2024unitime} have been integrated into time series, opening new avenues for zero-shot and cross-domain forecasting.


Despite these advancements, a critical challenge remains -- \textbf{\textit{the ability to identify and track concept drift}}, particularly in co-evolving time series where multiple series exhibit interdependent behavior over time. Concept drift -- changes in a series' statistical properties -- can significantly degrade model performance. This is crucial, particularly in fields like finance, where shifts in market regimes and nonlinear relationships are just as important as prediction accuracy for decision-makers. While recent methods like Cogra \cite{miyaguchi2019cogra} incorporate adaptive gradient techniques to address concept drift, they are limited to predefined concept structures and struggle with identifying transitions in dynamic, co-evolving environments. Similarly, Dish-TS \cite{fan2023dish} introduces distribution shift alleviation to improve forecasting performance. Notably, even the most recent deep learning methods that mention concept drift, such as OneNet \cite{wen2024onenet}, primarily aim to mitigate the impact of concept drift on forecasting rather than addressing the challenges of adaptive concept identification and dynamic concept drift.


Kolmogorov-Arnold Networks (KAN) \cite{liu2024kan} offer a promising solution to the challenges of concept drift in time series analysis. Inspired by the Kolmogorov-Arnold representation theorem, KAN replaces linear weights with spline-parametrized univariate functions, allowing the model to learn more complex relationships while improving both accuracy and interpretability. A notable advantage of KAN is its ability to refine spline grids, offering deeper insights into how inputs influence outputs, making the network’s decision-making process more transparent. However, while KAN has demonstrated strong performance with smaller network sizes across various tasks \cite{li2024u,shukla2024comprehensive}, its effectiveness in identifying and tracking concept drift within the time series domain remains unexplored.

To this end, our goal is to propose a KAN-based model for addressing concept drift in time series and evaluate its effectiveness. We introduce WormKAN, a concept-aware KAN-based model for co-evolving time series. Specifically, WormKAN processes co-evolving time series using (1) \textit{patch normalization} (PatchNorm), (2) \textit{kolmogorov-arnold self-representation networks} (KAN-SR), (3) \textit{temporal smoothness constraint} (TSC), (4) \textit{concept identification} (CI) and (5) \textit{concept drift} (CD) to uncover comprehensive concept transitions. PatchNorm treats co-evolving time series patches as the fundamental modeling units. KAN-SR leverages a KAN-based autoencoder with a novel self-representation layer to learn the representation and capture inter-patch dependencies. TSC applies a difference matrix constraint to the new representation, while CI and CD detect dynamic concept transitions. WormKAN identifies these transitions by detecting abrupt changes in the latent space, metaphorically described as \textbf{\textit{passing through a wormhole}}. These transitions mark shifts to new concepts, providing clear boundaries between different segments. Through experiments, we demonstrate that both the original KAN model and WormKAN effectively identify and track concept drift in time series. Our contributions are summarized as follows:
\begin{itemize}

\item[(1)] \textbf{Adaptive:} Automatically identify and handle concepts exhibited by co-evolving time series, without prior knowledge about concepts.

\item[(2)] \textbf{Interpretability:} Convert heavy sets of time series into a lighter and meaningful structure and depict the continuous concept shift mechanism.

\item[(3)] \textbf{Effective:} Operates on multiple time series patches, explores nonlinear interactions, and forecasts future concepts and values within a patch-based ecosystem.
\end{itemize}


\section{Related Work}
\subsection{Concept Drift in Time Series}
Concept drift presents a significant challenge in time series analysis, especially in streaming data environments where underlying data distributions evolve over time \cite{xu2024rhine,xu2024kernel,xu2018self,xu2022multi,xu2022data}. Traditional models like Hidden Markov Models (HMM) and Autoregression (AR) are commonly used but lack adaptability in continuous data streams. Recent advancements such as OrbitMap~\cite{matsubara2019dynamic} and Dish-TS~\cite{fan2023dish} have addressed scalability and distribution shifts but struggle with capturing temporal dependencies and dynamic transitions. Dynamic concept identification in co-evolving series is essential for understanding complex temporal patterns, with methods like AutoPlait~\cite{matsubara2014autoplait} providing segmentation capabilities. Deep learning techniques like OneNet~\cite{wen2024onenet} adapt to concept drift but prioritize predictive accuracy over interpretability. Concept-aware approaches, such as StreamScope~\cite{kawabata2019automatic}, attempt to capture behavioral transitions in temporal data. However, they often overlook interpretability, leaving a gap in providing clear demarcations of concept transitions for better understanding of dynamic temporal patterns.

\subsection{Kolmogorov-Arnold Networks (KAN)}
Kolmogorov-Arnold Networks (KAN) \cite{liu2024kan} leverage the Kolmogorov-Arnold representation theorem, which states that any multivariate continuous function can be represented using univariate functions. Specifically, a function \( f(x_1, x_2, \dots, x_n) \) can be expressed as \( f(x_1, x_2, \dots, x_n) = \sum_{q=1}^{2n+1} \Phi_q\left( \sum_{p=1}^{n} \varphi_{q,p}(x_p) \right) \), where $\varphi_{q,p}$ and $\Phi_q$ are univariate functions. Unlike MLPs, which rely on predefined activation functions at each node, KAN employs learnable activation functions along the network’s edges, offering enhanced flexibility and adaptability. This innovative design positions KAN as a compelling alternative to conventional MLPs for various applications. KAN replaces linear weights with learnable univariate functions, often parametrized using splines, allowing complex nonlinear relationships to be modeled with fewer parameters and greater interpretability. Inputs $x_p$ are transformed by $\varphi_{q,p}(x_p)$, aggregated, and passed through $\Phi_q$. Stacking multiple KAN layers allows the network to capture intricate patterns while maintaining interpretability, with the deeper architecture described as $\text{KAN}(x) = (\Phi_{L-1} \circ \cdots \circ \Phi_0)(x)$, where $L$ is the total number of layers.

KAN has been applied in various domains \cite{xukan4drift,xu2024kan,xu2024kolmogorov}. U-KAN \cite{li2024u} integrates KAN layers into the U-Net architecture, demonstrating impressive accuracy and efficiency in several medical image segmentation tasks. PIKAN \cite{shukla2024comprehensive} utilize KAN to build physics-informed machine learning models. This paper aims to introduce KAN to time series analysis and demonstrate its strong potential in representing time series data and effectively capturing concept drift.


\section{WormKAN}
\begin{figure}[!t]
\centerline{\includegraphics[width=0.8\linewidth]{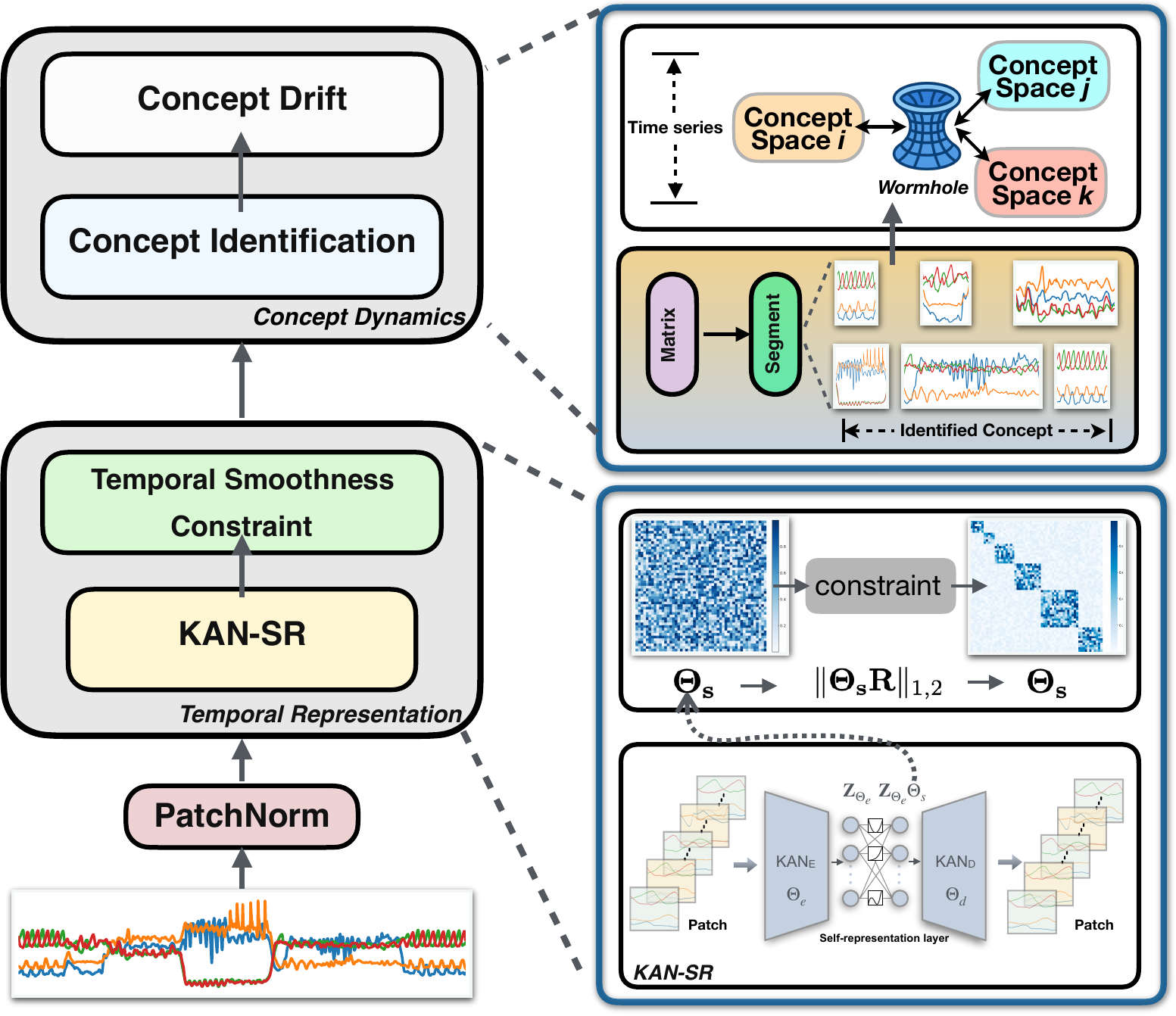}}
\caption{The framework of WormKAN. The co-evolving time series first passes through PatchNorm to transform into normalized patches, followed by a KAN-based Temporal Representation module to extract representations enriched with concept information. Finally, the Concept Dynamics module identifies concepts and tracks concept drift.}
\label{Fig1}

\end{figure}

Consider a co-evolving time series dataset $\mathbf{S} = \{S_1, S_2, \ldots,$  $S_N\} \in \mathbb{R}^{l \times N}$, where $N$ is the number of variables and $l$ represents the total number of time steps. Our goal is to (1) identify a set of latent concepts, $\mathbf{C} = {\mathbf{C}_1, \mathbf{C}_2, \ldots, \mathbf{C}_k}$, where $k$ is the number of concepts as they evolve; (2) track the evolution and drift over time; and (3) predict future concepts. Here, a concept reflects the joint behavior of the co-evolving time series, capturing their underlying interactions and temporal patterns. In this paper, we propose WormKAN, an architecture for concept-aware deep representation learning in co-evolving time series, as shown in Fig.~\ref{Fig1}. The architecture generally comprises three components: (1) \textit{patch normalization}, (2) a \textit{temporal representation}, and (3) \textit{concept dynamics}.


\subsection{Patch Normalization}
Unlike most existing methods, our approach models patches rather than individual points as the fundamental unit. Patches encapsulate more comprehensive information from local regions, offering richer representations compared to isolated points. Moreover, using patches reduces the sensitivity to inherent noise in the data, which can otherwise affect KAN-based models applied directly to individual points. This leads to more robust and stable representations.

To preserve the model’s auto-regressive nature, we set the patch length $w$ equal to the stride. This ensures that patches are non-overlapping segments of the original series, preventing access to future time steps and preserving the auto-regressive assumption. For simplicity, we assume the series length $l$ is divisible by $w$, yielding $n = l/w$ patches, denoted as $\mathbf{P} = { \mathbf{p}_1, \mathbf{p}_2, \ldots, \mathbf{p}_n }$, where each $\mathbf{p}_i$ spans multiple time steps and channels. This significantly reduces computational complexity, allowing the model to efficiently process longer series. Additionally, each patch undergoes instance normalization \cite{kim2021reversible} to standardize it with zero mean and unit variance. After predictions are made, the original mean and standard deviation are restored to maintain consistency in the final forecast.
\subsection{Temporal Representation}
Next, we apply the \textit{Temporal Representation} with the goal to learn a robust representation matrix, encapsulating the inherent dynamics of patches. Specifically, we process patches using (1) \textit{kolmogorov-arnold self-representation networks} (KAN-SR) and (2) \textit{temporal smoothness constraint}(TSC), which emphasizes the correlations within the patches.

\noindent \textbf{KAN-SR.} This module incorporates a KAN-based autoencoder combined with a self-representation layer. The autoencoder learns latent representations of input patches, while the self-representation layer captures relationships among these latent representations to ensure robust modeling of temporal dynamics. The encoder leverages a 3-layer KAN to map input patches $\mathbf{P}$ into a latent representation space through a nonlinear transformation: 
$\mathbf{Z}_{\Theta_e} = \text{KAN}_{\Theta_e}(\mathbf{P})$, where $\text{KAN}_{\Theta_e}$ denotes the encoding function, and $\mathbf{Z}_{\Theta_e}$ represents the resulting latent representations.

The self-representation layer identifies intrinsic relationships within the latent representations. Implemented as a 2-layer KAN with input and output layers, this layer enforces that each latent representation of patch can be expressed as a weighted combination of the others: $\mathbf{Z}_{\Theta_e} = \mathbf{Z}_{\Theta_e} \mathbf{\Theta_s}$, where $\mathbf{\Theta_s} \in \mathbb{R}^{n \times n}$ is the self-representation coefficient matrix. Each column $\bm{\theta}_{s,i}$ of $\mathbf{\Theta_s}$ represents the weights for reconstructing the $i$-th latent representation. To emphasize sparsity in $\mathbf{\Theta_s}$ and highlight significant relationships, we introduce an $\ell_1$ norm regularization: $\mathcal{L}_{\text{self}}(\mathbf{\Theta_s}) = \| \mathbf{\Theta_s} \|_1$.

The decoder reconstructs the input patches from the refined latent representations using another 3-layer KAN network: $\hat{\mathbf{P}}_{\Theta} = \text{KAN}_{\Theta_d}(\hat{\mathbf{Z}}_{\Theta_e})$, where $\text{KAN}_{\Theta_d}$ is the decoding function implemented using KAN, and $\hat{\mathbf{P}}_{\Theta}$ represents the reconstructed time series patches.

\noindent \textbf{Temporal Smoothness Constraint.} To ensure that the latent representations vary smoothly over time, we incorporate a temporal smoothness constraint on $\mathbf{\Theta_s}$. The core of this constraint is the difference matrix  $\mathbf{R} \in \mathbb{R}^{n \times (n-1)}$ , which captures the differences between consecutive columns of  $\mathbf{\Theta_s}$ . Specifically,  $\mathbf{R}$  is defined as:
\begin{equation}
\mathbf{R} =
\begin{bmatrix}
1 & -1 & 0 & \cdots & 0 \\
0 & 1 & -1 & \cdots & 0 \\
0 & 0 & 1 & \ddots & \vdots \\
\vdots & \vdots & \ddots & \ddots & -1 \\
0 & 0 & \cdots & 0 & 1 \\
\end{bmatrix}_{n \times (n-1)}
\end{equation}

The product $\mathbf{\Theta_s} \mathbf{R}$ captures the differences between consecutive columns: $\mathbf{\Theta_s} \mathbf{R} = [\bm{\theta}_{s,2} - \bm{\theta}_{s,1},\ \bm{\theta}_{s,3} - \bm{\theta}_{s,2},\ \ldots,\ \bm{\theta}_{s,n} - \bm{\theta}_{s,n-1}]$. The temporal smoothness constraint is defined as $\mathcal{L}_{\text{smooth}}(\mathbf{\Theta_s}) = \| \mathbf{\Theta_s} \mathbf{R} \|_{1,2}$, where $\| \cdot \|_{1,2}$ denotes the sum of the $\ell_2$ norms of the columns. This constraint penalizes large deviations, promoting smooth transitions and effectively capturing dynamic concept changes.

\noindent\textbf{Loss Function.} Training involves minimizing a loss function that combines reconstruction loss, self-representation regularization, and the temporal smoothness constraint:
\begin{equation}
\mathcal{L}(\Theta) = \tfrac{1}{2} \| \mathbf{P} - \hat{\mathbf{P}}_{\Theta} \|_F^2 + \lambda_1 \| \mathbf{\Theta_s} \|_1 + \lambda_2 \| \mathbf{Z}_{\Theta_e} - \mathbf{Z}_{\Theta_e} \mathbf{\Theta_s} \|_F^2 + \lambda_3 \| \mathbf{\Theta_s} \mathbf{R} \|_{1,2},
\end{equation}
where $\mathbf{\Theta} = \{ \mathbf{\Theta_e}, \mathbf{\Theta_s}, \mathbf{\Theta_d} \}$ includes all learnable parameters, with $\lambda_1$, $\lambda_2$, and $\lambda_3$ balancing the different loss components. Specifically, $\lambda_1$ promotes sparsity in the self-representation $\Theta_s$, $\lambda_2$ preserves the self-representation property by minimizing the difference between $\mathbf{Z}_{\Theta_e}$ and $\mathbf{Z}_{\Theta_e} \mathbf{\Theta_s}$, and $\lambda_3$ ensures temporal smoothness by reducing deviations in the temporal difference matrix $\mathbf{\Theta_s} \mathbf{R}$.

\subsection{Concept Dynamics}

\begin{figure}
  \centering
  \includegraphics[width=0.7\textwidth]{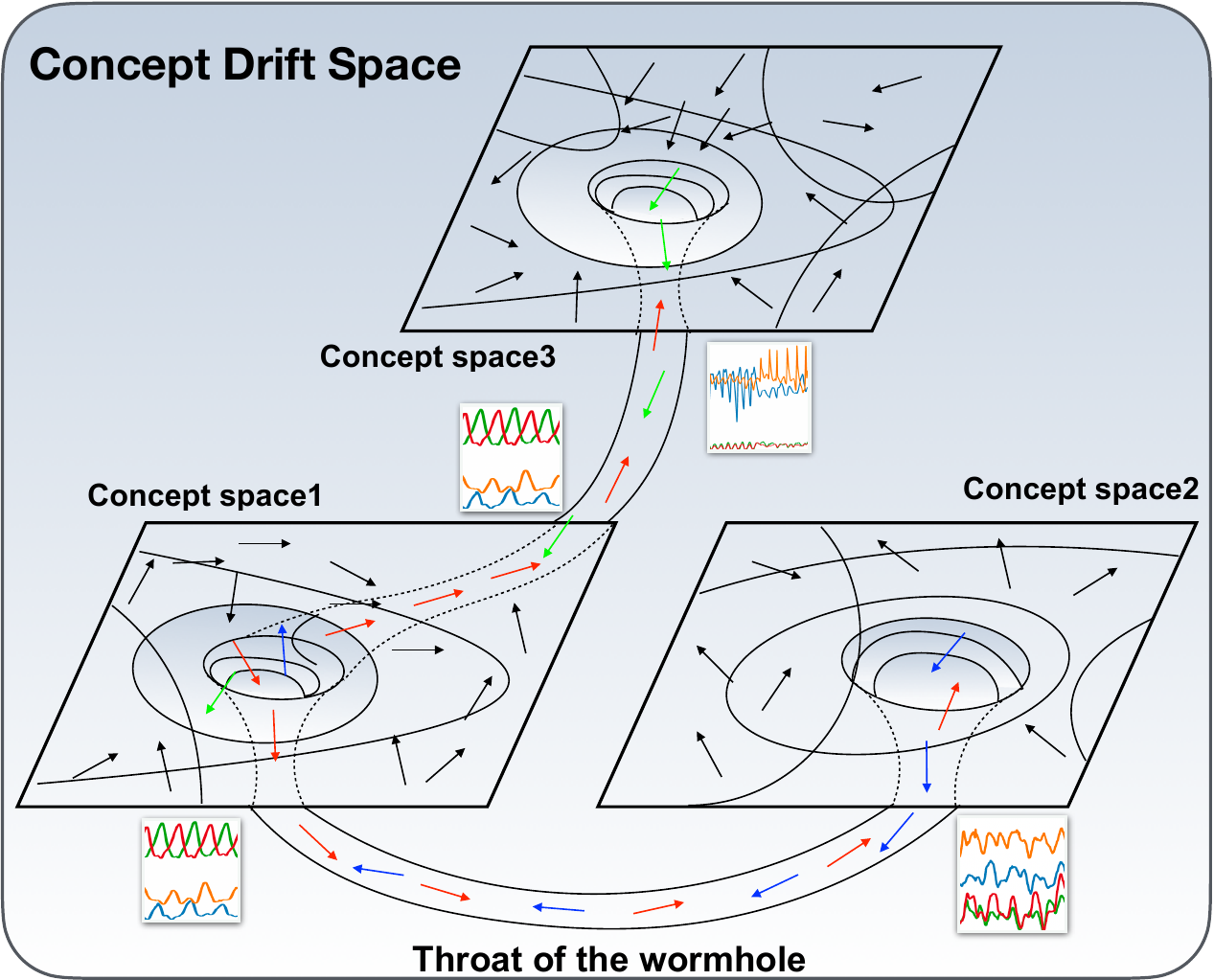}
\caption{Concept transition through \textit{wormhole}.}\label{Fig_worm}
\vspace{-10pt}
\end{figure}
Concept Dynamics explores the transitions and evolution of distinct concepts. These transitions are identified through the analysis of the representation matrix $\mathbf{\Theta_s} \mathbf{R}$, capturing both concept boundaries and dynamic changes over time. We describe this process in terms of two complementary aspects: \textit{Concept Identification} and \textit{Concept Drift}.

\noindent \textbf{Concept Identification.}
This aspect focuses on segmenting the self-representation matrix $\mathbf{\Theta_s}$ to uncover distinct concept regions. Transitions between concepts are identified by analyzing the deviations in $\mathbf{\Theta_s}\mathbf{R}$, where significant changes signal the boundaries of segments. To achieve segmentation, we compute the absolute value matrix $\mathbf{B} = |\mathbf{\Theta_s} \mathbf{R}|$ and derive the column-wise mean vector $\mu^{\mathbf{B}}$. A peak-finding algorithm is then applied to $\mu^{\mathbf{B}}$, with peaks indicating segment boundaries. This process enables the identification of concept regions, each representing a unique latent behavior within the time series.

\noindent \textbf{Concept Drift.}
Concept drift captures the temporal dynamics of transitions between distinct concept regions. Beyond detecting segment boundaries by monitoring deviations in $\mathbf{\Theta_s} \mathbf{R}$, each identified concept is associated with a prototype -- a patch whose vector representation aligns with the centroid of similar patches within the same concept space. At any time, if the patch begins to approach this prototype, it signals the likelihood of an imminent concept drift. These transitions, metaphorically described as \textbf{passing through wormholes}, signify abrupt shifts in behavior and structure between concept spaces. Crossing a segment boundary indicates departing from the current concept space and entering a new one, while approaching the prototype suggests being drawn into the gravitational pull of the new concept region. These transitions effectively highlight dynamic changes in the structure of the time series, enabling the model to detect, capture, and adapt to evolving behaviors (Figure~\ref{Fig_worm}).

\subsection{Forecasting}

Although forecasting is not the primary focus of this work, we provide a simple yet effective autoregressive method to predict future concepts and, based on the predicted concepts, forecast subsequent series values. Specifically, the future concept $\mathbf{C}_{k+1}$ is predicted as:

\begin{equation}
    \mathbf{C}_{k+1}=\lambda(\mathbf{C}_1,..., \mathbf{C}_k)+\mu_{k+1},
\label{predictC}
\end{equation}
where $\lambda(\cdot)$ represents an autoregressive function over the historical concepts, and $\mu_{k+1}$ is white Gaussian noise added to reduce overfitting.

Based on the predicted concept $\mathbf{C}_{k+1}$, we forecast the value of the time series by leveraging a representation derived from historical patches belonging to the same concept. Specifically, the forecast is expressed as:
\begin{equation}
\hat{\mathbf{p}}_{n+1} = \sum_{i \in \mathcal{H}(\mathbf{C}_{k+1})} \alpha_i \mathbf{p}_i,
\label{predictX}
\end{equation}
where $\mathcal{H}(\mathbf{C}_{k+1})$ represents the set of historical patches associated with the concept $\mathbf{C}_{k+1}$, $\alpha_i$ are the autoregressive weights, and $\mathbf{p}_i$ are the historical patch values.

\section{Property of the Self-representation Layer $\mathbf{\Theta_s}$}
The core of WormKAN is the representation matrix $\mathbf{\Theta_s}$, which encapsulates the relationships and concepts within time series patches. Without loss of generality, let $\mathbf{P} = [\mathbf{P}^{(1)}, \mathbf{P}^{(2)}, \cdots, \mathbf{P}^{(k)}]$ be ordered according to their concept. Ideally, we wish to obtain a representation $\mathbf{\Theta_s}$ such that each patch is represented as a combination of points belonging to the same concept, i.e., $\mathbf{P}^{(i)} = \mathbf{P}^{(i)}\mathbf{\Theta_s}^{(i)}$. In this case, $\mathbf{\Theta_s}$ has the $k$-block diagonal structure, i.e.,

\begin{equation}
\begin{aligned}
\mathbf{\Theta_s}=&
\begin{bmatrix}

\mathbf{\Theta_s}^{(1)} &0& \cdots & 0 \\

0&\mathbf{\Theta_s}^{(2)} &\cdots&0\\

\vdots &\vdots & \ddots & \vdots \\

0 &0& \cdots & \mathbf{\Theta_s}^{(k)}

\end{bmatrix}
\end{aligned}
\label{eq:sr}
\end{equation}

This representation reveals the underlying structure of $\mathbf{P}$, with each block $\mathbf{\Theta_s}^{(i)}$ in the diagonal representing a specific concept. $k$ represents the number of blocks, which is directly associated with the number of distinct concept.

\section{Experiments}

In this section, we first evaluate the original KAN model's effectiveness in detecting concept drift, followed by experiments validating the performance of WormKAN for concept-aware deep representation learning in co-evolving time series.

\subsection{Data}
To evaluate the original KAN, we use the synthetic \texttt{SyD} dataset with 500 co-evolving time series, providing controllability over concept structures and ground truth for evaluation. For WormKAN’s ability to identify meaningful concepts, we employ three datasets: Motion Capture Streaming Data (MoCap) from the CMU database\footnote{\url{http://mocap.cs.cmu.edu/}}, which records transitions between activities; Stock Market data with historical prices and financial indicators from 503 companies\footnote{\url{https://ca.finance.yahoo.com/}}; and Online Activity Logs from Google Trends\footnote{\url{http://www.google.com/trends/}}, consisting of 20 queries from 2004 to 2022. For forecasting, we use the publicly available ETT, Traffic, and Weather datasets. The Electricity Transformer Temperature (ETT) datasets\footnote{\url{https://github.com/zhouhaoyi/ETDataset}} include four subsets: \texttt{ETTh1}, \texttt{ETTh2} (hourly-level) and \texttt{ETTm1}, \texttt{ETTm2} (15-minute-level), with seven features recorded from 2016 to 2018. The \texttt{Traffic} dataset\footnote{\url{http://pems.dot.ca.gov}} provides hourly road occupancy rates across 862 sensors on San Francisco freeways from 2015 to 2016. The \texttt{Weather} dataset\footnote{\url{https://www.bgc-jena.mpg.de/wetter/}} contains 21 meteorological indicators recorded every 10 minutes in Germany during 2020.
\subsection{Original KAN Model Evaluation}
\begin{figure}[t]
\centerline{\includegraphics[width=0.9\linewidth]{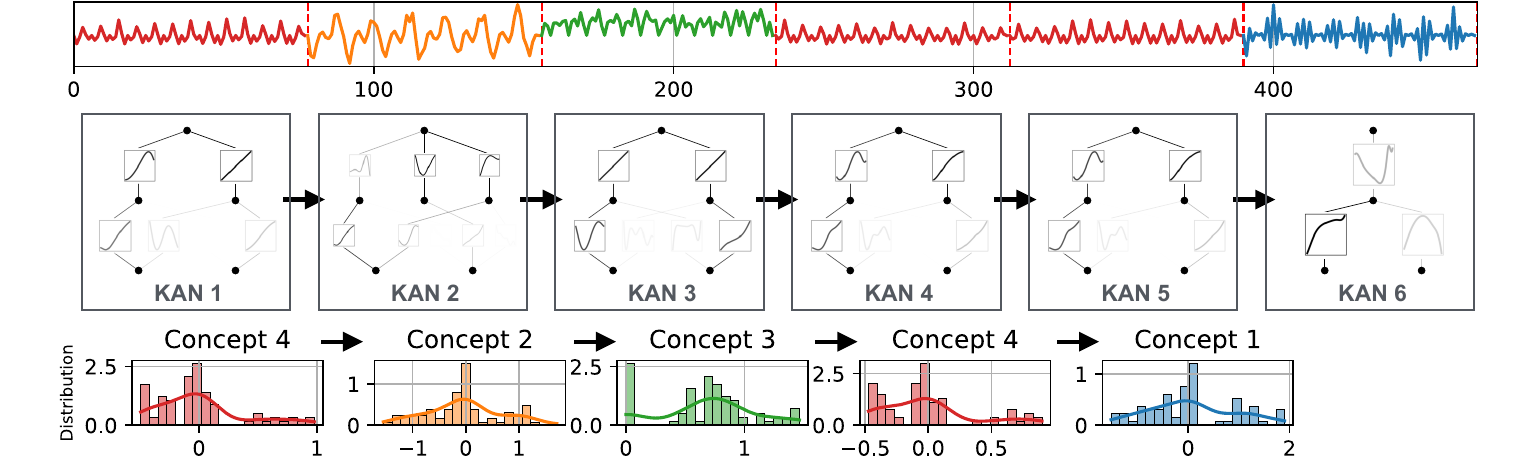}}

\caption{Training original KAN and detecting concept drift.}
\label{Fig2}

\end{figure}

As illustrated in Figure~\ref{Fig2}, we use a sliding window to traverse the synthetic time series, creating input-output pairs for training the KAN model. For simplicity, two historical time steps predict the next step. Different KAN structures and activation functions represent different concepts, and observing variations in KAN can reveal concept drift. For example, the learnable activation functions in KAN1, KAN4, and KAN5 behave consistently, indicating they are within the same concept.

\subsection{WormKAN Performance and Baseline Comparison}

To evaluate WormKAN, we assess its performance from two perspectives: (1) its ability to identify meaningful concepts within co-evolving time series (Concept Dynamics) and (2) its effectiveness in time series forecasting tasks (Forecasting).

\noindent \textbf{Concept Dynamics.} We evaluated WormKAN’s ability to identify important concepts and transitions using three co-evolving time series datasets: human motion, financial markets, and online activity logs. WormKAN was compared against StreamScope \cite{kawabata2019automatic}, TICC \cite{hallac2017toeplitz}, and AutoPlait \cite{matsubara2014autoplait}, which are methods for discovering concepts in co-evolving series. The results in Table \ref{tab} show that WormKAN outperforms all baselines.

\begin{figure}[!t]
\centerline{\includegraphics[width=0.8\linewidth]{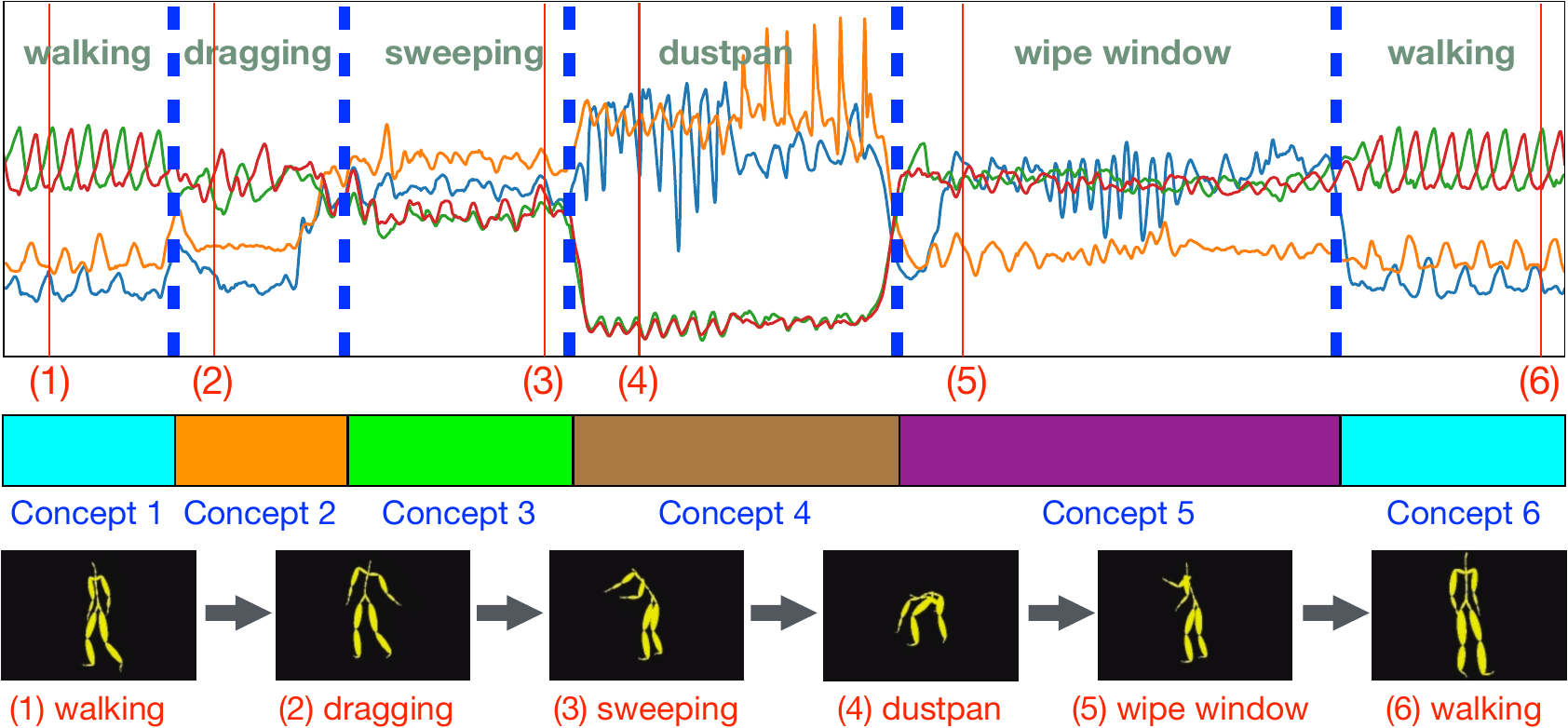}}
\caption{WormKAN identifies concepts and transitions on co-evolving motion series.}
\label{Fig3}

\end{figure}





\begin{table} 
   \centering
\caption{Comparison with Baseline Models}
\label{tab}
\resizebox{0.7\textwidth}{!}
{
\begin{tabular}{|c|c|c|c|c|c|}
\hline
Dataset & Metric & StreamScope & TICC & AutoPlait & WormKAN \\ \hline

\multirow{2}{*}{Motion Capture Data} & F1-Score & 0.84 & 0.48 & 0.87 & \textbf{0.90} \\ \cline{2-6}
 & ARI & 0.60 & 0.22 & 0.60 & \textbf{0.65} \\ \hline

\multirow{2}{*}{Stock Market Data} & F1-Score & 0.75 & 0.32 & 0.80 & \textbf{0.86} \\ \cline{2-6}
 & ARI & 0.62 & 0.20 & 0.74 & \textbf{0.82} \\ \hline

\multirow{2}{*}{Online Activity Logs} & F1-Score & 0.92 & 0.80 & 0.90 & \textbf{0.94} \\ \cline{2-6}
 & ARI & 0.85 & 0.75 & 0.83 & \textbf{0.90} \\ \hline
\end{tabular}}

\end{table}

\begin{table*}[!th]

\caption{Models' forecasting performance, in terms of RMSE. The best results are highlighted in \textcolor{red}{\textbf{bold}} and the second best are \textcolor{blue}{\underline{underlined}}.}
\label{Tab:performanc_real}
\centering
\resizebox{\textwidth}{!}{
\begin{tabular}{l c cccc cccc|c}
\toprule
\multirow{2}{*}{\textbf{Datasets}} & \multirow{2}{*}{\textbf{Horizon}}  & \multicolumn{4}{c}{\textbf{Forecasting models}} & \multicolumn{5}{c}{\textbf{Concept-aware models}} \\ 
\cmidrule(lr){3-6} \cmidrule(lr){7-11}
 &  & ARIMA & KNNR & INFORMER & N-BEATS & Cogra & OneNet & OrbitMap & \textbf{KAN} & \textbf{WormKAN} \\
\midrule
\multirow{4}{*}{\texttt{ETTh1}} 
                                 & 96  & 1.209 & 0.997 & 0.966 & \textcolor{blue}{\underline{0.933}} & 0.948 &  \textcolor{red}{\pmb{0.916}} & 0.945 & 2.009 & \textcolor{blue}{\underline{0.933}} \\
                                 & 192 & 1.267 & 1.034 & 1.005 & 1.023 & 0.996 & \textcolor{blue}{\underline{0.986}} & 0.991 & 2.154 & \textcolor{red}{\pmb{0.975}} \\
                                 & 336 & 1.297 & 1.057 & \textcolor{blue}{\underline{1.035}} & 1.048 & 1.041 & \textcolor{red}{\pmb{1.028}} & 1.039 & 2.240 & \textcolor{red}{\pmb{1.028}} \\
                                 & 720 & 1.347 & 1.108 & 1.088 & 1.115 & \textcolor{blue}{\underline{1.080}} & 1.082 & 1.083 & 2.361 & \textcolor{red}{\pmb{1.079}} \\
\midrule
\multirow{4}{*}{\texttt{ETTh2}} 
                                 & 96  & 1.216 & 0.944 & 0.943 & \textcolor{blue}{\underline{0.892}} & 0.901 & \textcolor{blue}{\underline{0.892}} & 0.894 & 1.947 & \textcolor{red}{\pmb{0.889}} \\
                                 & 192 & 1.250 & 1.027 & 1.015 & \textcolor{blue}{\underline{0.979}} & 0.987 & \textcolor{red}{\pmb{0.968}} & 0.976 & 2.149 & 0.983 \\
                                 & 336 & 1.335 & 1.111 & 1.088 & \textcolor{red}{\pmb{1.040}} & 1.065 & \textcolor{blue}{\underline{1.049}} & 1.052 & 2.297 & \textcolor{red}{\pmb{1.040}} \\
                                 & 720 & 1.410 & 1.210 & 1.146 & \textcolor{red}{\pmb{1.101}} & 1.131 & 1.120 & 1.120 & 2.453 & \textcolor{blue}{\underline{1.119}} \\
\midrule
\multirow{4}{*}{\texttt{ETTm1}} 
                                 & 96  & 0.997 & 0.841 & 0.853 & 0.806 & 0.780 & \textcolor{blue}{\underline{0.777}} & \textcolor{red}{\pmb{0.765}} & 1.718 & \textcolor{red}{\pmb{0.765}} \\
                                 & 192 & 1.088 & 0.898 & 0.898 & 0.827 & 0.819 & 0.813 & \textcolor{red}{\pmb{0.810}} & 1.771 & \textcolor{blue}{\underline{0.812}} \\
                                 & 336 & 1.025 & 0.886 & 0.885 & 0.852 & 0.838 & 0.823 & \textcolor{blue}{\underline{0.820}} & 1.808 & \textcolor{red}{\pmb{0.819}} \\
                                 & 720 & 1.070 & 0.921 & 0.910 & 0.903 & 0.890 & \textcolor{red}{\pmb{0.859}} & \textcolor{blue}{\underline{0.868}} & 1.901 & 0.874 \\
\midrule
\multirow{4}{*}{\texttt{ETTm2}} 
                                 & 96  & 0.999 & 0.820 & \textcolor{blue}{\underline{0.812}} & \textcolor{red}{\pmb{0.804}} & 0.824 & 0.822 & 0.821 & 1.782 & \textcolor{red}{\pmb{0.804}} \\
                                 & 192 & 1.072 & 0.874 & 0.902 & 0.839 & 0.849 & 0.835 & \textcolor{blue}{\underline{0.832}} & 1.815 &\textcolor{red}{\pmb{0.829}} \\
                                 & 336 & 1.117 & 0.905 & \textcolor{blue}{\underline{0.842}} & 0.852 & 0.854 & 0.846 & 0.852 & 1.863 & \textcolor{red}{\pmb{0.839}} \\
                                 & 720 & 1.176 & 0.963 & 0.965 & \textcolor{red}{\pmb{0.890}} & 0.921 & \textcolor{blue}{\underline{0.896}} & 0.906 & 1.949 & \textcolor{blue}{\underline{0.896}} \\
\midrule
\multirow{4}{*}{\texttt{Traffic}} 
                                 & 96  & 1.243 & 1.006 & 0.895 & 0.893 & 0.898 & \textcolor{blue}{\underline{0.887}} & 0.889 & 1.936 & \textcolor{red}{\pmb{0.883}} \\
                                 & 192 & 1.253 & 1.021 & 0.910 & 0.920 & 0.908 & 0.913 & \textcolor{blue}{\underline{0.905}} & 1.954 & \textcolor{red}{\pmb{0.898}} \\
                                 & 336 & 1.260 & 1.028 & 0.916 & 0.909  & 0.922 & \textcolor{blue}{\underline{0.904}} & 0.908 & 2.061 & \textcolor{red}{\pmb{0.901}} \\
                                 & 720 & 1.285 & 1.060 & 0.968 & 0.949 & 0.964 & \textcolor{red}{\pmb{0.940}} & \textcolor{blue}{\underline{0.946}} & 2.050 & \textcolor{red}{\pmb{0.940}} \\
\midrule
\multirow{4}{*}{\texttt{Weather}} 
                                 & 96  & 1.013 & 0.814 & 0.800 & 0.752 & 0.759 &\textcolor{blue}{\underline{0.747}} & \textcolor{red}{\pmb{0.744}} & 1.621 & \textcolor{red}{\pmb{0.744}} \\
                                 & 192 & 1.021 & 0.867 & 0.861 & 0.798 & 0.793 & \textcolor{blue}{\underline{0.783}} & 0.789 & 1.696 & \textcolor{red}{\pmb{0.776}} \\
                                 & 336 & 1.043 & 0.872 & 0.865 & 0.828 & 0.825 & \textcolor{blue}{\underline{0.811}} & 0.816 & 1.740 & \textcolor{red}{\pmb{0.801}} \\
                                 & 720 & 1.096 & 0.917 & 0.938 & \textcolor{red}{\pmb{0.833}} & 0.863 & 0.853 & \textcolor{blue}{\underline{0.841}} & 1.848 & \textcolor{red}{\pmb{0.833}} \\
\bottomrule
\end{tabular}}
\parbox{\textwidth}{\footnotesize\textcolor{magenta}{While forecasting series task is not our main focus, we provide a comparison of original KAN and WormKAN with other models. }}

\end{table*}   

We also visualized the concept transitions detected by WormKAN on the Motion Capture Data. Figure~\ref{Fig3} illustrates time series segments with marked transitions between different motion types, such as walking and dragging. These visualizations highlight WormKAN's ability to accurately detect boundaries between different types of activity, reinforcing its effectiveness in identifying dynamic changes in co-evolving series.

\noindent \textbf{Forecasting.} For real-world datasets, we lack the ground truth for validating the obtained concepts. Instead, we validate the value and gain of the discovered concepts for time series forecasting. We evaluate the forecasting performance of the proposed model against seven different models, utilizing the Root Mean Square Error (RMSE) as an evaluative metric. These seven models include four forecasting models (ARIMA \cite{box2013box}, KNNR \cite{box2013box}, INFORMER \cite{zhou2021informer}, and a ensemble model N-BEATS \cite{oreshkin2019n}), and three are concept-aware models (Cogra \cite{miyaguchi2019cogra}, OneNet \cite{wen2024onenet} and OrBitMap \cite{matsubara2019dynamic}).

Table~\ref{Tab:performanc_real} shows the forecasting performance of the models. We observe that our model, WormKAN, consistently outperforms other models, achieving the lowest forecasting error on most datasets. \textbf{Notably, the original KAN exhibits poor forecasting performance, likely due to its simplicity, as it consists of only the original two layers (we did not include additional layers)}. N-BEATS, a state-of-the-art deep learning model, performs effectively due to its ensemble-based architecture. However, it does not consistently perform as well as WormKAN or OneNet, particularly in handling concept drift across multiple time series. OneNet, similar to N-BEATS, achieves strong results owing to its ensemble-based strengths. On datasets where variables/channels are relatively independent and concept drift follows predictable patterns, WormKAN’s results are not as competitive as those of OneNet, which benefits from its ensemble forecasting approach. (Note that ensemble learning models generally outperform single models.) Nevertheless, WormKAN, as a single-model framework, achieves comparable outcomes to OneNet.

\begin{figure}[t]
\centerline{\includegraphics[width=0.8\linewidth]{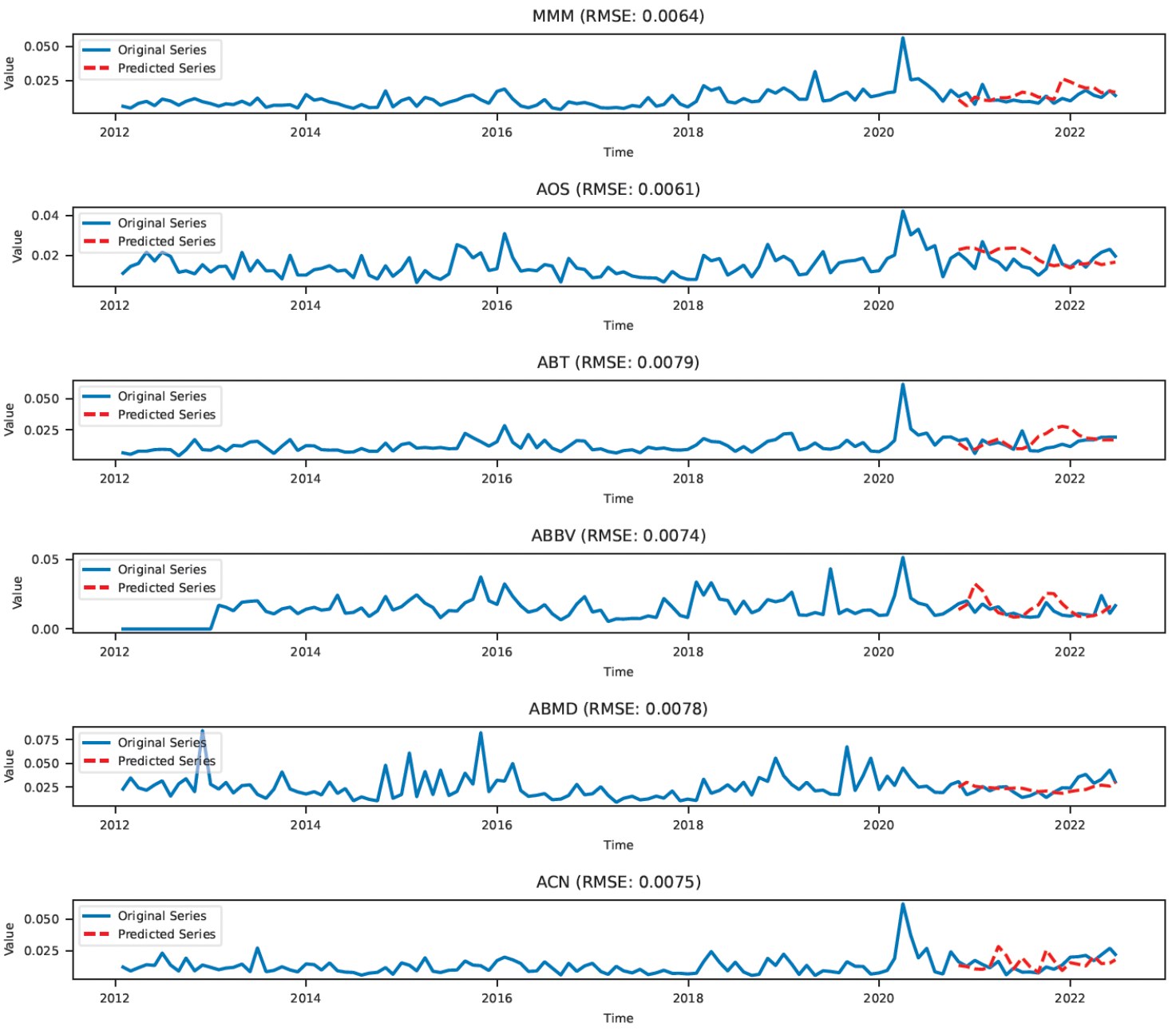}}

\caption{Predicted results using WormKAN: true (blue) and forecasted (red) values for the first six stock series.}
\label{Fig4}

\end{figure}

Furthermore, we conducted additional experiments to explore WormKAN’s forecasting ability on the Stock dataset. We visualized the original series (in blue) and the predicted results (in red) for the first six stocks (Nasdaq: MMM, AOS, ABT, ABBV, ABMD, ACN) from the dataset. As shown in Figure~\ref{Fig4}, these visualizations highlight the notable efficacy of our model in forecasting time series, offering deeper insights into its practical performance.

\section{Conclusion}

This work demonstrates the effectiveness of Kolmogorov-Arnold Networks (KANs) in detecting concept drift in time series. We introduced WormKAN, a concept-aware KAN-based model for co-evolving time series. By leveraging patch normalization, we construct co-evolving time series as patches to fundamental modeling units. With kolmogorov-arnold self-representation networks and temporal smoothness constraint, WormKAN learns the robust representation and captures inter-patch dependencies. Through concept identification and drift, WormKAN identifies dynamic concept transitions. Our results highlight KANs' potential for robust, adaptive modeling in dynamic time series environments.

\bibliographystyle{elsarticle-num}  
\bibliography{references}

\end{document}